\documentclass[10pt,twocolumn,letterpaper]{article}

\usepackage{cvpr}              

\usepackage{graphicx}
\usepackage{amsmath}
\usepackage{amssymb}
\usepackage{booktabs}
\usepackage{enumerate}
\usepackage{multirow}

\usepackage[pagebackref,breaklinks,colorlinks]{hyperref}

\usepackage[capitalize]{cleveref}
\crefname{section}{Sec.}{Secs.}
\Crefname{section}{Section}{Sections}
\Crefname{table}{Table}{Tables}
\crefname{table}{Tab.}{Tabs.}

\def\cvprPaperID{46} 
\def\confName{CVPR}
\def\confYear{2022}

\begin{document}

\title{Spatial-temporal Concept based Explanation of 3D ConvNets}

\author{Ying Ji\\
Nagoya University, Japan\\
\and
Yu Wang\\
Hitotsubashi University, Japan\\
\and
Kensaku Mori\\
Nagoya University, Japan\\
\and
Jien Kato\\
Ritsumeikan University, Japan\\
}

\maketitle

\begin{abstract}
Recent studies have achieved outstanding success in explaining
2D image recognition ConvNets. On the other hand, due to the computation cost and complexity of video data, the explanation of 3D video recognition ConvNets is relatively less studied. In this paper, we present a 3D ACE (Automatic Concept-based Explanation) framework for interpreting 3D ConvNets. In our approach: (1) videos are represented using high-level supervoxels, which is straightforward for human to understand; and (2) the interpreting
framework estimates a score for each voxel, which reflects its importance in the decision procedure. Experiments show that our method can discover spatial-temporal concepts of different importance-levels, and thus can explore the influence of the concepts on a target task, such as action classification, in-depth. The codes are publicly available \footnote{\url{https://github.com/OrangeeJi/3D-ACE}}.
\end{abstract}

\section{Introduction}
\label{intro}

Deep ConvNets have shown remarkable performance on various tasks. Despite the popularity, their decision procedure still lacks transparency and interpretability. Recently, high-level concepts have been utilized in explaining 2D image recognition ConvNets. Kim \textit{et al.}~\cite{kim2018interpretability} introduced concept activation vectors (CAVs) which use the directional derivatives to quantify the importance of user-defined concepts in 2D ConvNets. Based on \cite{kim2018interpretability}, ~\cite{ghorbani2019towards} further proposed ACE (Automatic Concept-based) to explore the relationship between meaningful image segments and class predictions.

On the other hand, only few studies tried to interpret 3D video recognition ConvNets, mainly due to the huge computation cost and high complexity of video data. Limited existing works mainly rely on the recurrent neural networks (RNN) to generate temporal attention, which leads to two main disadvantages: (1) spatial and temporal attention are separately generated, (2) the extracted attention segments are pixel-level regions, lacking human-understandable high-level information.

To address these issues, we extend 2D ACE \cite{ghorbani2019towards} to 3D ACE and propose to use spatial-temporal concepts to interpret the decision procedure of 3D video recognition ConvNets. We validate our method on the Kinetics dataset using popular network architectures and visualize the results. Both qualitative and quantitative results show that our approach can interpret the 3D ConvNets consistent with human cognition.

\section{Proposed method}
\label{sec:method}
Given a video classification dataset and a 3D ConvNet trained from it, we interpret the later one by investigating the most important spatial-temporal volumes from the earlier one. Videos are first segmented into supervoxels. Similar supervoxels within each class are then clustered and lead to a set of spatial-temporal concepts. 3D ACE finally evaluates the importance score of each concept with respect to the class it belongs. Within the decision procedure, the network pays more attention to the concepts with high score.

\subsection{Supervoxel representation} 
Let $V = {\left\{(v_n, y_n)\right\}}_{n=1}^N$ be a video classification dataset of $N$ videos, where $v_n$ is the $n$th video and $y_n\in (1, Y)$ is the label. Each video is segmented $3$ times with different levels of resolution, which can preserve hierarchical information. $[s_n^{small}, s_n^{middle}, s_n^{large}]$ contains different size of supervoxels for video $v_n$. To avoid computational cost for redundant supervoxels, we calculate the similarity between every two supervoxels. Only the most distinguishable supervoxels will remain. A scratched 3D ConvNet is trained on $V$ and used as a feature extractor. The feature vector is extracted from the top layer for each supervoxel.

\subsection{Concept-based explanation} 
After extracting deep features, supervoxels in class $y$ are clustered into $C$ concepts. $s_c^y$ denotes all the segments belongs to the $c$th concept. 

We then evaluate the importance score for each concept. First, a concept activation vector (CAV) \cite{kim2018interpretability} is calculated to characterize the concept. $s_c^y$ are put into the ConvNet as positive samples, while a group of random videos is set as negative samples. A linear classifier is learned to separate the positive and negative samples. The vector $v_c^l$ that orthogonal to the decision boundary is used to represent $c$.

To figure out the influence of the concept $c$ given to a video $v_n$ when ConvNet predicting, we follow the idea from \cite{kim2018interpretability} to calculate the gradient of logit with respect to the activations of $v_n$ in layer $l$. Thus the importance score of a certain concept can be computed as $I_{c,y,l}(v_n)$:

\begin{eqnarray}    \label{eq1}
I_{c,y,l}(v_n) &=& \lim_{\epsilon \to 0} \frac{p_{l,y}(f_l(v_n)+ \epsilon v_c^l)-p_{l,y}(f_l(v_n))}{\epsilon}\nonumber    \\
~&=& \nabla p_{l,y}(f_l(v_n)) \cdot v_c^l
\end{eqnarray}
where $f_l(v_n)$ is the feature vector of the input video, and $p_{l,y}$ is the logit for the video $v_n$ from class $y$. 

For one class with $K$ input videos, the total importance score of one concept is defined as: 

\begin{eqnarray}
S_{c,y,l}  &=& \frac{\left| {v_n\in V:  I_{c,y,l}(v_n)} > 0 \right|}{K} \in [0,1]
\end{eqnarray}

For concept $c$, the score $S_{c,y,l}$ computes the fraction of input videos that are positively influenced by the concept. And the higher $S$ indicates the most concerning part for a ConvNet to recognize the video.

\begin{figure}
  \centering
  \begin{subfigure}{0.48\linewidth}
    \includegraphics[width = \textwidth]{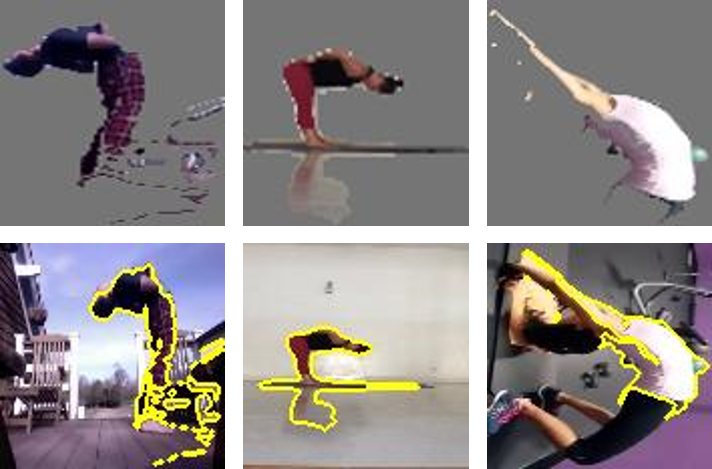}
    \caption{}
    \label{fig:most}
  \end{subfigure}
  \hfill
  \begin{subfigure}{0.48\linewidth}
    \includegraphics[width = \textwidth]{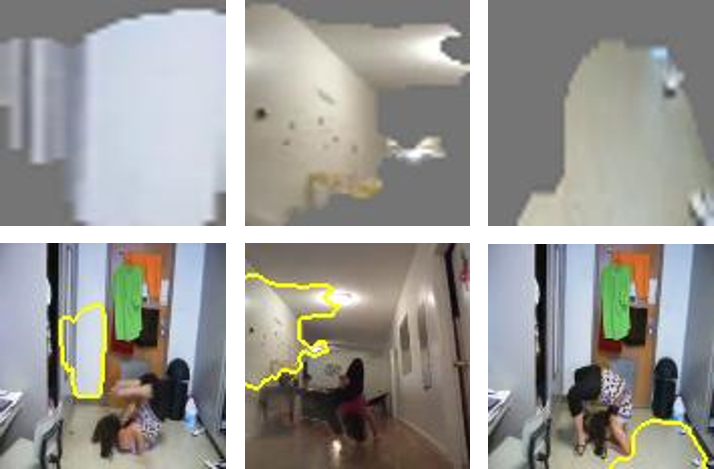}
    \caption{}
    \label{fig:least}
  \end{subfigure}
  \caption{Example of the most and the least important concepts.}
  \label{fig:all}
 \vspace*{-4mm}
\end{figure}

\section{Experiment}
\label{exp}
In this section, we present experimental results on the proposed 3D ACE. We evaluate the performance by removing and adding concepts from the video dataset and visualizing the concepts. $10$ classes randomly selected from Kinetics-700 \cite{carreira2019short} are uesd to conduct the experiment. 

Table \ref{table:add} shows the results of adding concepts using different ConvNets. For each model, the first row is adding the highest score concepts. The second row is adding concepts randomly. The third row is adding the lowest score concepts. We observe that adding the most important concepts can lead to a higher accuracy. After adding 5 concepts, the accuracy can reach over $70\%$ of the baseline. Table \ref{table:remove} demonstrates the influence of removing concepts. It can be seen that removing the most important concepts will obviously decrease the accuracy. These results indicate we successfully reveal which concept the 3D ConvNets focus on and how much role it plays during the prediction.

We also visualize the concepts extracted with 3D ACE from the ``bending back" class. Figure \ref{fig:most} shows the concepts with the highest importance scores. It can be observed that the highlighted regions mainly focus on the bending action and the body part. Figure \ref{fig:least} shows the lowest score concepts. By contrast, these regions are almost meaningless.

\begin{table}[h]
\caption{The result of adding concepts. The baseline is the classification accuracy ($\%$) on the chosen $10$ classes dataset.}
\label{table:add}
\resizebox{0.49\textwidth}{!}{
\begin{tabular}{cclllllc}
\hline
Model                & Concepts & \multicolumn{1}{c}{1} & \multicolumn{1}{c}{2} & \multicolumn{1}{c}{3} & \multicolumn{1}{c}{4} & \multicolumn{1}{c}{5} & Baseline                 \\ \hline
\multirow{3}{*}{r3d~\cite{tran2018closer}} & Top      & 11.67               & 23.96              & 32.92               & 39.38               & \textbf{46.67}      & \multirow{3}{*}{75.62} \\
                     & Random   & 10.63               & 21.25              & 32.50               & 37.71              & 41.25               &                          \\
                     & Least    & 9.79               & 16.04               & 26.04               & 33.13               & 41.46              &                          \\ \hline
\multirow{3}{*}{i3d~\cite{carreira2017quo}} & Top      & 23.33               & 37.92               & 46.88               & 54.38               & \textbf{61.88}      & \multirow{3}{*}{85.63} \\
                     & Random    & 25.83               & 37.50               & 46.04               & 52.71              & 56.67               &                          \\
                     & Least   & 25.42               & 37.50               & 47.29               & 51.46               & 55.83               &                          \\ \hline
\end{tabular}}
\vspace*{-5mm}
\end{table}

\begin{table}[h]
\caption{Classification accuracy $(\%)$ of removing concepts.}
\label{table:remove}
\resizebox{0.49\textwidth}{!}{
\begin{tabular}{cclllllc}
\hline
Model                & Concepts & \multicolumn{1}{c}{1} & \multicolumn{1}{c}{2} & \multicolumn{1}{c}{3} & \multicolumn{1}{c}{4} & \multicolumn{1}{c}{5} & Baseline                 \\ \hline
\multirow{3}{*}{r3d~\cite{tran2018closer}} & Top      & 69.79               & 66.25               & 50.83               & 39.58               & \textbf{24.38}      & \multirow{3}{*}{75.62} \\
                     & Random   & 72.29\              & 64.38               & 51.04               & 39.79               & 28.13               &                          \\
                     & Least    & 73.33               & 64.38               & 51.67               & 42.29              & 28.13               &                          \\ \hline
\multirow{3}{*}{i3d~\cite{carreira2017quo}} & Top      & 74.58               & 65.63              & 58.96               & 49.79               & \textbf{41.88}      & \multirow{3}{*}{85.63} \\
                     & Random   & 78.33              & 70.83              & 60.42             & 53.75               & 43.96               &                          \\
                     & Least    & 80.21               & 71.25               & 65.00               & 57.92              & 46.25               &                          \\ \hline
\end{tabular}}
\vspace*{-5mm}
\end{table}

\section{Conclusion}
\label{con}
In this paper, we proposed a spatial-temporal concept-based explanation framework for 3D ConvNets. Different from previous low-level pixel-based method, our research provide a human-understandable high-level explanation. Extensive experiments show the effectiveness of our proposed method. 

{\small
\bibliographystyle{ieee_fullname}
\bibliography{reference}
}

\end{document}